\documentclass{article}

\usepackage{arxiv}

\usepackage[utf8]{inputenc} %
\usepackage[T1]{fontenc}    %
\usepackage{hyperref}       %
\usepackage{url}            %
\usepackage{booktabs}       %
\usepackage{amsfonts}       %
\usepackage{nicefrac}       %
\usepackage{microtype}      %
\usepackage{lipsum}

\usepackage{graphicx}
\usepackage{float}
\usepackage{algorithm2e} 

\usepackage{color,soul}

\makeatletter
\newif\if@anonymize

\@anonymizefalse  %

\if@anonymize
\usepackage{tikz}
\usetikzlibrary{calc}
\usepackage{censor,caption}
\newcommand{\anonymizeemail}[1]{none@anonymize.invalid}
\newcommand{\anonymizeurl}[1]{http://www.xxx.xxx/}
\else
  \newcommand{\xblackout}[1]{#1}
  
  \newcommand{\anonymizeemail}[1]{#1}
  \newcommand{\anonymizeurl}[1]{#1}
\fi
\makeatother

\usepackage{color}
\usepackage{xcolor}
\usepackage{ifthen}
\let\oldcite=\cite
\renewcommand\cite[1]{\ifthenelse{\equal{#1}{NEEDED}}{\colorbox{red}{\textbf{[??]}}}{\oldcite{#1}}}

\usepackage{balance}

\hypersetup{
pdftitle={Group-Level Emotion Recognition Using a Unimodal Privacy-Safe Non-Individual Approach},
pdfsubject={Computer vision, Neural networks},
pdfauthor={Anastasia Petrova, Dominique Vaufreydaz, and Philippe Dessus},
pdfkeywords={EmotiW 2020, audio-video group emotion recognition, Deep Learning, affective computing, privacy},
}

\renewcommand{\today}{\vspace{-1em}}

\title{Group-Level Emotion Recognition Using a Unimodal Privacy-Safe Non-Individual Approach}

\author{
	Anastasia Petrova\\
    Univ. Grenoble Alpes, CNRS, Inria, Grenoble INP, LIG, 38000 Grenoble, France\\
	\texttt{Anastasia.Petrova@inria.fr} 
	\And
	Dominique Vaufreydaz\\
    Univ. Grenoble Alpes, CNRS, Inria, Grenoble INP, LIG, 38000 Grenoble, France\\
	\texttt{Dominique.Vaufreydaz@univ-grenoble-alpes.fr}
	\And
	Philippe Dessus\\
    Univ. Grenoble Alpes, LaRAC, 38000 Grenoble, France\\
	\texttt{Philippe.Dessus@univ-grenoble-alpes.fr}
}

\newcommand{\teachinglab}{\textit{\xblackout{Teaching Lab}}}

\begin{document}

\maketitle

\begin{abstract}

This article presents our unimodal privacy-safe and non-individual proposal for the audio-video group emotion recognition subtask at the Emotion Recognition in the Wild (EmotiW) Challenge 2020\footnote{\href{https://sites.google.com/view/emotiw2020/}{https://sites.google.com/view/emotiw2020/}, last accessed 08/2020.}.
This sub challenge aims to classify \textit{in the wild} videos into three categories: Positive, Neutral and Negative. Recent deep learning models have shown tremendous advances in analyzing interactions between people, predicting human behavior and affective evaluation. Nonetheless, their  performance comes from individual-based analysis, which means summing up and averaging scores from individual detections, which inevitably leads to some privacy issues. In this research, we investigated a frugal approach towards a model able to capture the global moods from the whole image without using face or pose detection, or any individual-based feature as input.
The proposed methodology mixes state-of-the-art and dedicated synthetic corpora as training sources.
With an in-depth exploration of neural network architectures for group-level emotion recognition, we built a VGG-based model achieving 59.13\% accuracy on the VGAF test set (eleventh place of the challenge). Given that the analysis is unimodal based only on global features and that the performance is evaluated on a real-world dataset, these results are promising and let us envision extending this model to multimodality for classroom ambiance evaluation, our final target application.

\end{abstract}


\section{Introduction}

Since the emergence of the “\textit{affective computing}” research trend at the beginning of the 2000s~\cite{picard2003affective}, more and more features describing people's affective states are calculable using machine learning on remote sensor signals.
Deep learning has attracted significant attention of the research community thanks to its faculty to automatically extract features from images and videos, determining the best representations from raw data. Deep learning shows high performance in multiple domains, notably in computer vision and speech recognition. The literature presents a wide variety of Convolutional Neural Networks (CNNs) algorithms with extensive applications in many areas, such as human-computer interaction, image retrieval, surveillance, virtual reality, etc. These achievements impact other research areas, such as affective computing and behavioral modeling to detect and predict human actions, being one of the main multimodal perception instruments for analyzing interactions between people and groups~\cite{barsade2015group}.  However, most of them cannot reach the human-level understanding of \textit{in the wild} emotions \cite{10.1371/journal.pone.0231968}.

In the \teachinglab{} project, we aim at developing a smart pervasive classroom to analyze teacher--students social relationships from still images or video sequences coupled with acoustic features: namely current teacher activity, current teaching episodes, whole-class engagement, students' attention or engagement, classroom ambiance, etc. The system's goal is to analyze the underlying teaching processes (e.g., teacher–students interaction, misbehavior management), not to monitor individual behaviors \textit{per se}, even if they are inadequate. The multimodal perception system will thus monitor the whole classroom at a glance to help teachers enhance their pedagogical practices afterward. The underlying research question is: is it possible to use deep learning to capture global moods only from the whole scene?

While perceiving groups, current state-of-the-art perception systems 
focus on people as individuals, i.e., each individual is detected and processed as one entity. Group level perception is then computed by iterating over the group. 
For instance, to detect which emotion is carried by a photograph, systems typically try to locate faces accurately and then identify emotions from facial expressions with techniques such as facial action coding. Averaging over results leads to the final estimation~\cite{dhall2016emotiw}. However, this approach can raise problems for privacy and ethical concerns. An ambient classroom has “ears and eyes” and any attendee's behavior is subject to be recorded and further analyzed~\cite{ahuja2019edusense,laurent:hal-02438020} for monitoring or surveillance purposes, thus hampering their so-called "\textit{ambient privacy}".
Since the overall goal of \teachinglab{} systems is to obtain global information about pedagogical episodes, a question arises about the possibility to calculate it without individual sub-calculations. This kind of approach exists using audio processing~\cite{james2019automated}, but, as far as we know, no research on pure global group-level emotions features on images/videos, again without including individual computations, is reported in the literature.

This paper addresses our first investigation about perceiving the mood of groups of people with an ethical and privacy-safe frugal approach, i.e. an ethical approach excluding individual-based features. The EmotiW 2020 challenge \cite{emotiw2020} and its Audio-video Group Emotion Recognition sub-challenge are an opportunity for the challenge team (\textit{OnlyGlobalFeatures}) to confront this problem using data gathered ``\textit{in the wild}'' and sustain its ethical proposal using deep neural network using only global features. The current proposed unimodal model achieved 59.13\% of accuracy on the VGAF test set~\cite{sharma2019automatic} and the eleventh place in the challenge. Using only the video stream, this promising preliminary result is an opening on future evolutions of such perception systems and will be further extended to audio signal processing.

This article is organized as follows. The related work section provides insights into current researches on group-level perception. Then, the proposed approach and the system pipeline are depicted. The next sections describe information about the training datasets and conducted experiments. Results are discussed and conclusions are drawn in the last sections. 

\section{Related work}

A significant contribution to automatic computer recognition of human emotions was made by the previous EmotiW challenges participants. Audio-video Group Emotion Recognition is a new problem introduced this year, aiming to detect and classify the emotion of a group of people among three classes: positive, neutral and negative. This new task is the combination of two sub-challenges that were held in previous years. Despite the considerable progress that has been made in the field of computer vision, predicting the emotions of a group of people in the wild is still a difficult task, due to the number of faces in the image, various illumination conditions, face deformations, occlusions, non-frontal faces,  etc. For this challenge, the organizers proposed a baseline based on the Inception-V3 architecture~\cite{szegedy2016rethinking}. Their work includes the analysis of visual and audio cues~\cite{sharma2019automatic}. 

One of these sub-challenges is Group-level emotion recognition (GReco) 2018, which consists in predicting the mood of a group of people from images. Participants combined the top-down (the usage of the global tone of a photo for analysis) and bottom-up (performing analysis for each of the extracted faces, and then the subsequent combination of the results) approaches to design their prediction algorithm. The winner proposed a hybrid deep learning network~\cite{guo2018group} obtained by fusing four individually trained models for faces, scenes, skeletons and salient regions detection using VGGNet~\cite{simonyan2014very}, ResNet~\cite{he2016deep}, Inception~\cite{szegedy2015going} and LSTM~\cite{hochreiter1997long} networks. The adjustment of each model's prediction weights by grid-search strategy allowed achieving a 68.08\% classification accuracy on the testing set. The second-place system proposed a cascade attention network for the face cues and CNNs (VGG19~\cite{simonyan2014very}, ResNet101~\cite{he2016deep}, SE-Net154~\cite{hu2018squeeze}) for body, face and image cues \cite{wang2018cascade}. Participants used the cascade attention network to solve the partial discrepancy of some faces to the photograph's general atmosphere, which allowed them to capture the importance of each face in the image and achieve 67.48\% classification accuracy. For face and body detection, the researchers used MTCNN~\cite{zhang2016joint} and OPENPOSE~\cite{cao2017realtime}, respectively. The third-place proposed a four-stream hybrid network, consisting of the face-location aware global stream, the multi-scale face stream, the global blurred stream and the global stream, which achieved 65.59\% accuracy on the
test set \cite{khan2018group}. For face detection, this team also used MTCNN~\cite{zhang2016joint}. The face stream consisted of two networks designed for high and low-resolution face images. To learn features from the whole image, they fine-tuned a pre-trained VGG16~\cite{simonyan2014very} network on the ImageNet dataset~\cite{deng2009imagenet}. The face-location aware global stream was specially implemented to capture features from the whole image together with face-relative location information. To learn only scene related features, the participants proposed a global blurred stream, where original images with blurred faces were used as input. Finally, all four streams were fused by tuning fusion weights.

The second sub-challenge, Audio-video Emotion Recognition task (VReco), aims to predict six basic emotions plus neutral from videos, was organized seven times since 2013. In the 2019 EmotiW challenge, Li et al. \cite{li2019bi} proposed a bimodality fusion framework containing two models. The first one is the facial image model, where four different types of neural networks (VGG-Face, Restnet18, Densenet121, VGG16) were used for feature extraction, and the obtained features were integrated into a Bi-Directional Long Short Term Memory (Bi-LSTM) model~\cite{schuster1997bidirectional} to capture dynamical temporal information. The second one is the audio model based on two methods for feature extraction and
classification, using deep learning feature extraction and the openSMILE toolbox \cite{eyben2010opensmile} to compute LLD features. The fusion of the two models achieved 62.78\% accuracy on the test set and ranked 2nd. Zhou et al. \cite{zhou2019exploring} explored emotion features (revealing the most suitable CNNs) and three types of fusion strategies (self-attention, relation-attention, and transformer), that are usually used to highlight the essential emotional features.
Their model obtained 62.48\% accuracy and ranked at the 3rd place.

\begin{figure*}[ht!]
  \centering
  \includegraphics[width=0.9\textwidth]{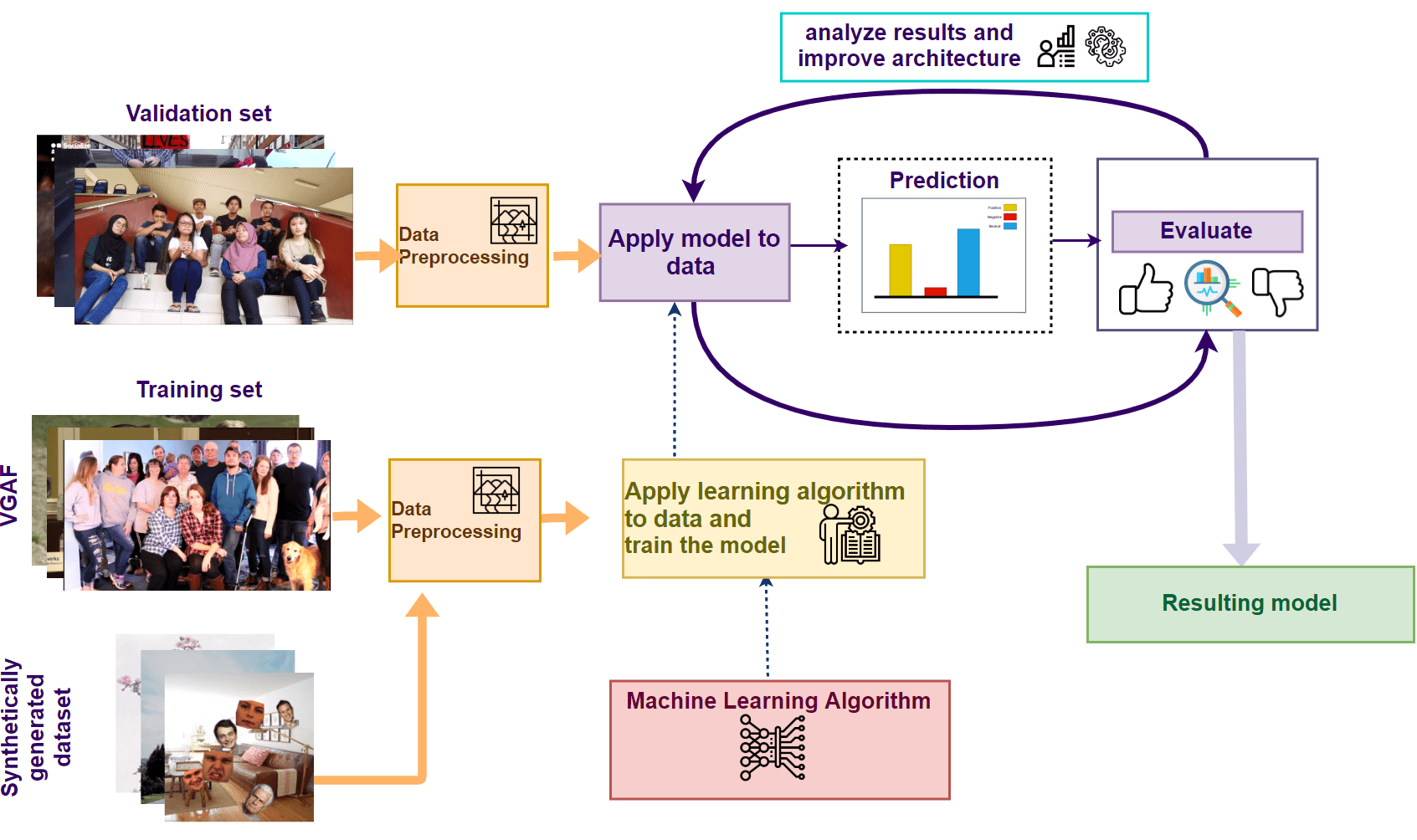}
  \caption{The pipeline for the proposed approach to group emotion recognition. This scheme reflects the key points of our analysis, namely the process of model selection, training and adjustment.}
  \label{fig:schemeofapproach}
\end{figure*}

\section{Methodology} %

As already stated, this research voluntarily positions itself in a frugal approach excluding the prior detection of faces or the body poses, and more generally, the detection of individuals. It differs from usual approaches that combine global features with individual ones.
As traditional CNNs are by nature not explainable, i.e. one does not know what has actually been learned, our global proposal takes advantage of this fact to increase the privacy of monitored users.
Recent studies use individualized visual cues coupled with these global visual features in classrooms, some also combining them with global acoustic features \cite{khan2018group,guo2018group,wang2018cascade,ramakrishnan2019toward}.
In this first investigation about pure global group mood recognition, we will focus only on visual information, trying to reach the best performance before mixing with audio information. This study and our participation in the challenge is the first step towards a more comprehensive classroom ambiance recognition.

That said, as depicted in Figure~\ref{fig:schemeofapproach}, our methodology explores different convolutional neural network architectures, focusing those that have shown superior results for computer vision tasks over the past years.
These systems were trained on the training data and compared by evaluating their score on the challenge's validation set, revealing the most suitable architecture to such a global approach.
We then selected the best initial model and gradually modified and refined it to increase the prediction accuracy on the validation set. The rationale is to assess the baseline performance of a global approach. 

The available data to train group emotion recognition systems is limited. We decided to augment it with synthetic data generated on-the-fly during the training. This allows us to add a variety in terms of facial expressions, positions of faces and background types. This generation mechanism is described in section~\ref{sec:synthdataset}.

\section{Datasets}

This section presents the EmotiW 2020 dataset for the Audio-video Group Emotion Recognition sub-challenge and our synthetic dataset generation dedicated to group mood detection within images.

\subsection{The VGAF dataset}

The VGAF dataset~\cite{sharma2019automatic} is the set of videos for group-level emotion recognition collected \textit{in the wild}. It contains 1,004 videos collected from the Web with different real-world scenarios, faces, illuminations, occlusions. The data is distributed into three parts: Train, Validation, and Test with 587, 215 and 202 samples respectively. These videos were collected from different social events, such as protests, TV shows, interviews,  parties, meetings, sport events, etc. There are three categories in this dataset: Positive, Negative, Neutral. The duration of each video is between 8 and 25 seconds. The number of people present in each video ranges from 2 to 62. Data annotation was done manually by three different persons. The final label has been agreed upon by all annotators.

\begin{figure}[h]
  \centering\hfill
\includegraphics[width=0.3\textwidth]{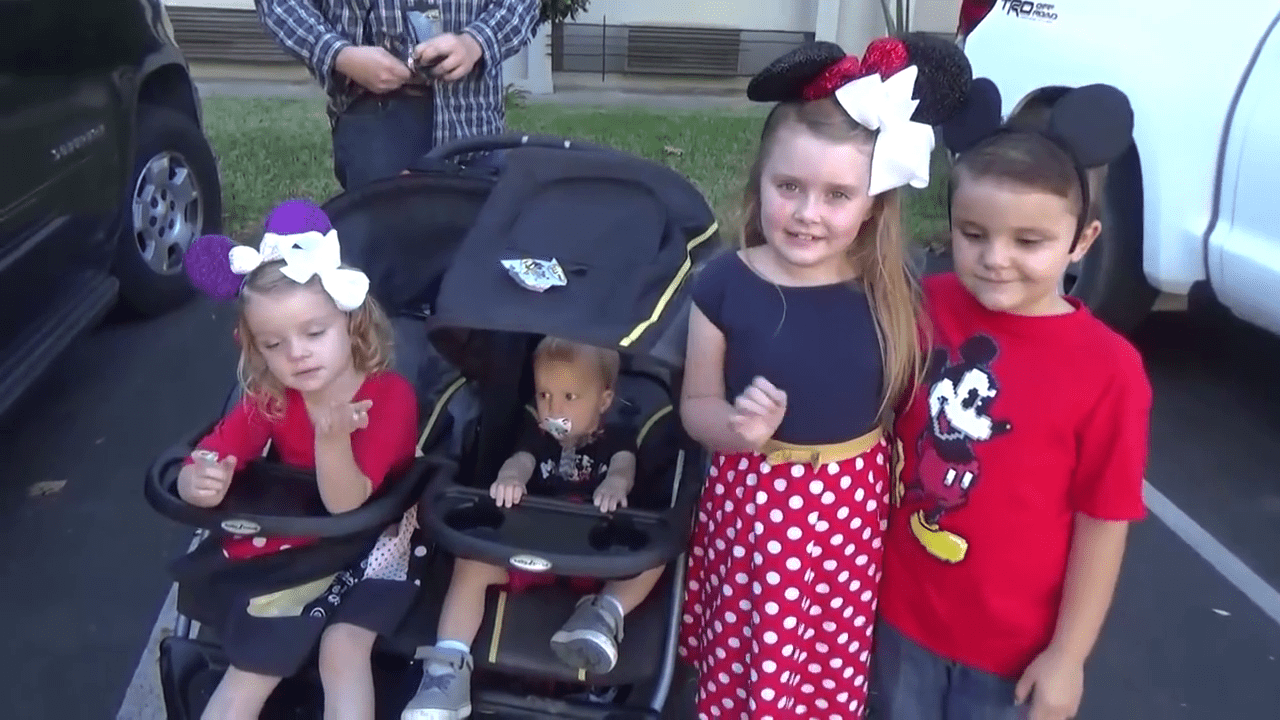}\hfill
\includegraphics[width=0.3\textwidth]{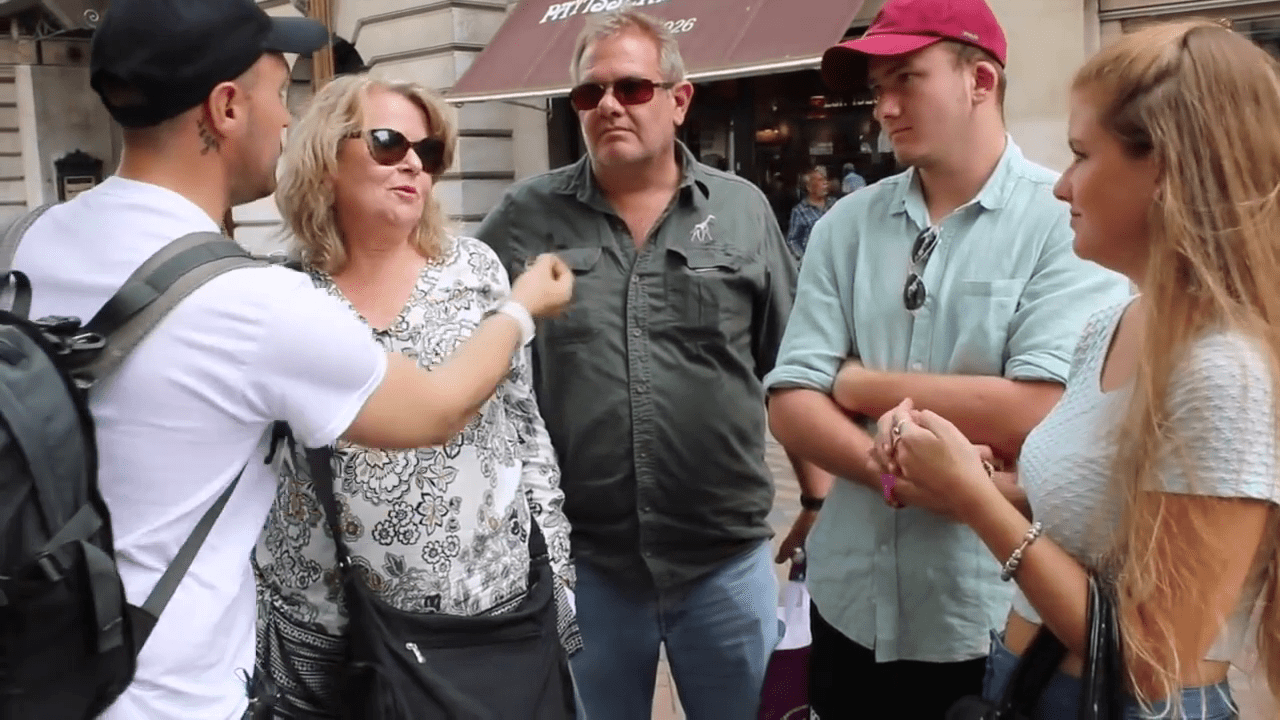}\hfill
\includegraphics[width=0.3\textwidth]{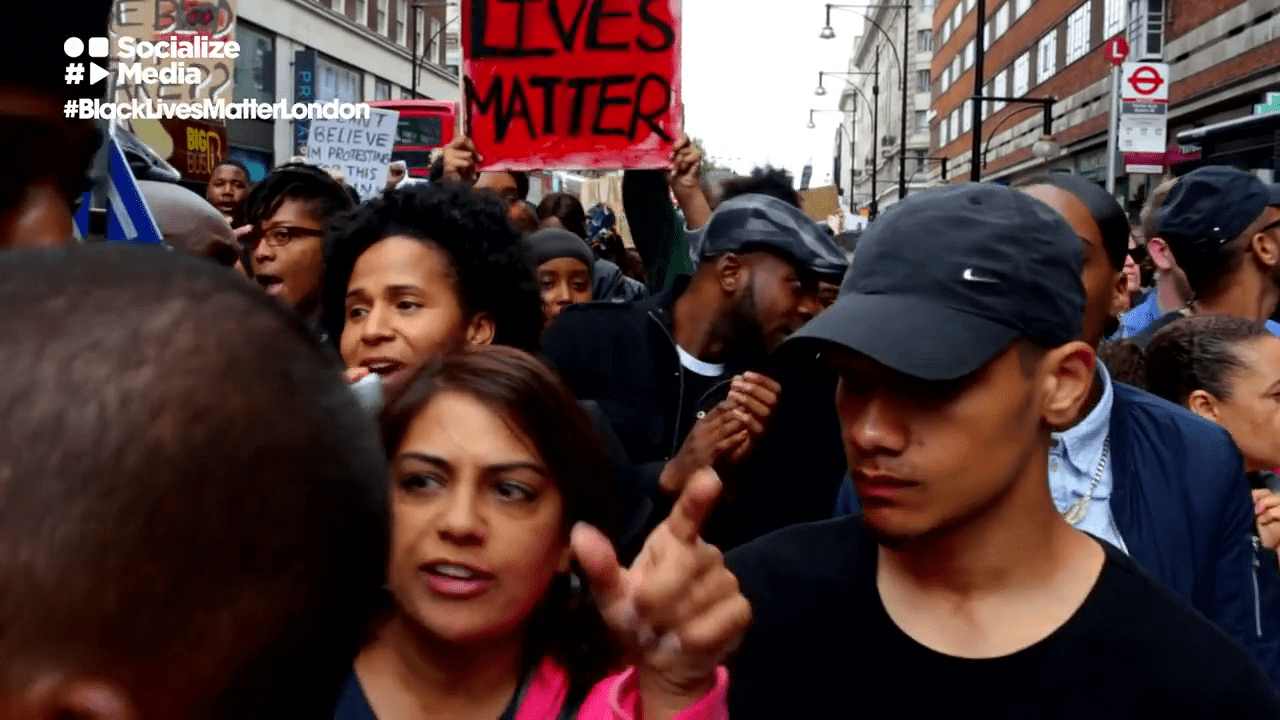}\hfill
  \caption{Examples of frames from VGAF \protect\cite{sharma2019automatic}  dataset, from left to right: Positive, Neutral and Negative images. Three frames from VGAF dataset for Positive, Neutral and Negative classes, respectively. }
  \label{fig:VGAF}
\end{figure}

Figure~\ref{fig:VGAF} highlights the difficulty of the task through examples of extracted frames from the VGAF~\cite{sharma2019automatic} train dataset for each class. In most cases, the images contain non-frontal faces, occlusions, not clear backgrounds. The images contain spontaneous emotions that people express in everyday life.

\subsection{Synthetic dataset}\label{sec:synthdataset}

Besides VGAF, we created a synthetically generated dataset to enlarge the number of training samples. We used an approach similar to those described in \cite{varol2017learning}. The process consists in creating images using real faces showing different emotions and superimpose them on an arbitrary background. The underlying idea is to guide the neural network while training to focus on faces in images while ignoring background pixels.

\subsubsection{Generation algorithm}
\begin{figure}
    \centering
\includegraphics[width=0.8\linewidth]{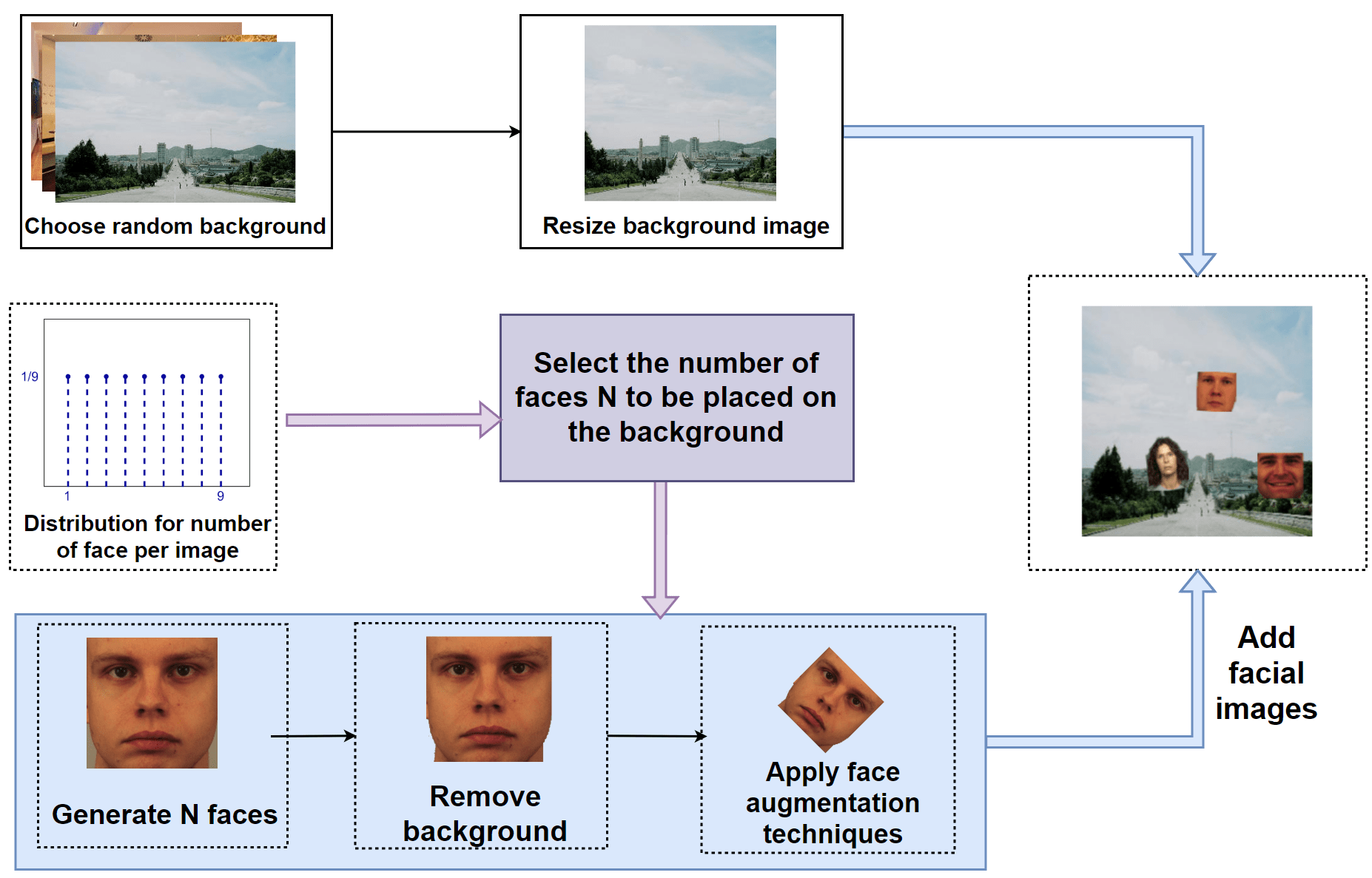}
\caption{Synthetic Dataset Generation Scheme. This scheme represents the key steps of data generation process}
\label{fig:datageneration}
\end{figure}

The generation process is depicted in Figure~\ref{fig:datageneration}. It consists of several key steps. First, the generator selects a background and \textit{N} faces randomly. Analyzing statistics from the VGAF dataset, the number of people present on each video ranges from 2 to 62, while the mode equals 9. Most videos contain 4 to 18 faces. Knowing that, we decided to set $N \in [1,9]$.
After data augmentation on selected faces, the generation algorithm places them on the image at random positions but, using a binary mask, the algorithm prevents occlusions between faces. Then, it computes the image label.
All negative emotions (\textit{anger}, \textit{fear}, \textit{disgust}, \textit{sadness}) were combined into the class '\textit{Negative}',  \textit{happiness} and \textit{neutral} respectively into  '\textit{Positive}' and '\textit{Neutral}' classes. Each of the generated images contains emotions from different classes. The emotion label is set as the most common class presented in the image. If none, it is set to \textit{neutral}. Figure~\ref{fig:syntheticdata} shows examples of generated images for each class. The next sections detail these generation steps.

\subsubsection{Background images}
\begin{figure}
    \centering
\hfill
\includegraphics[width=0.3\textwidth]{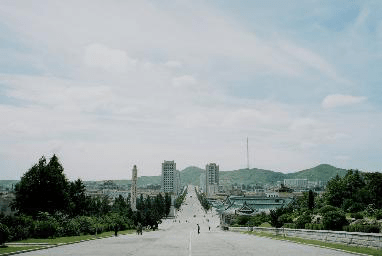}\hfill
\includegraphics[width=0.3\textwidth]{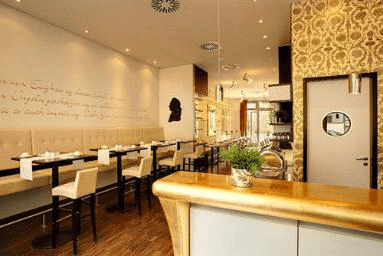}\hfill
\includegraphics[width=0.3\textwidth]{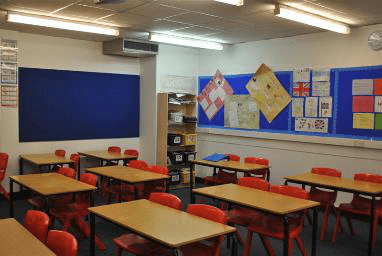}\hfill
\caption{Background images from LSUN \protect\cite{yu2015lsun} dataset, from left to right: tower, restaurant, classroom. Three examples of background images used for synthetic data generation.}
\label{fig:LSUN}
\end{figure}

The first step in creating synthetic images is the background selection. The goal of the generation process is to feed the neural network with many different environments to learn to ignore them: extracting only emotions, regardless of background. Our expectation is that the system trained with such images would learn to focus on people.
The dataset LSUN~\cite{yu2015lsun} is a good background provider as it contains 10 different scene categories: classroom, bedroom, bridge, church outdoor, conference room, dining room, kitchen, living room, restaurant, tower. Figure \ref{fig:LSUN} shows some examples of background images.

\subsubsection{Face images augmentations}
\begin{figure}
\centering
\includegraphics[width=0.09\textwidth]{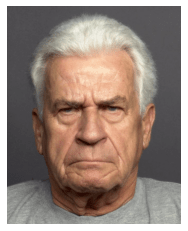}\includegraphics[width=0.09\textwidth]{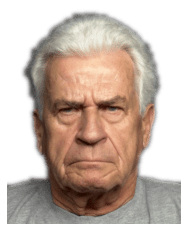}\includegraphics[width=0.09\textwidth]{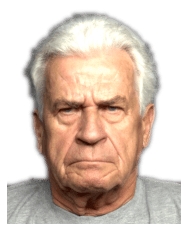}\includegraphics[width=0.09\textwidth]{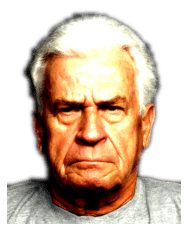} \includegraphics[width=0.09\textwidth]{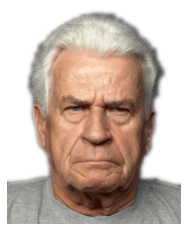} \includegraphics[width=0.09\textwidth]{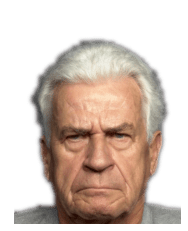}\includegraphics[width=0.09\textwidth]{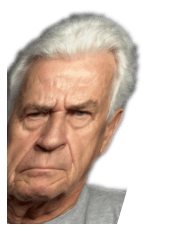}\includegraphics[width=0.09\textwidth]{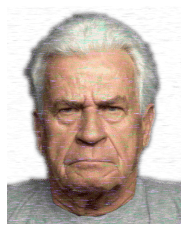}\includegraphics[width=0.09\textwidth]{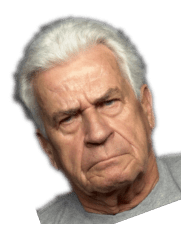}\includegraphics[width=0.09\textwidth]{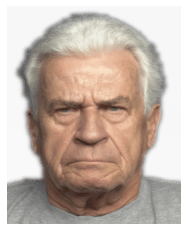}\includegraphics[width=0.09\textwidth]{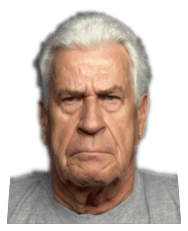}\\
\caption{Examples of face augmentations, from top to bottom, from left to right: source image from \protect\cite{ebner2010faces}, source image without background, brightness change, contrast change, horizontal flip, translate, shear, elastic distortion, rotate, scale, perspective transformation. Examples of data transformations used for facial images.}
\label{fig:Data_transformations}
\vspace{2em}\hfill
\includegraphics[width=0.3\textwidth]{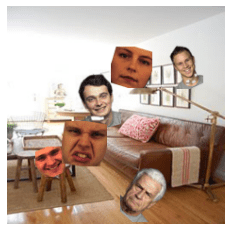}\hfill
\includegraphics[width=0.3\textwidth]{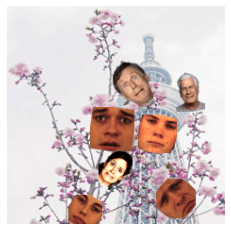}\hfill
\includegraphics[width=0.3\textwidth]{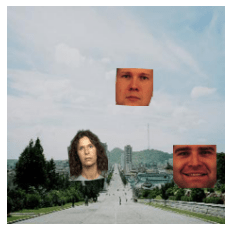}\hfill
\caption{Examples of synthetically generated data. Labels from left to right: Positive, Negative, Neutral. These images shows three examples of generated data: facial images on the arbitrary background.}
\label{fig:syntheticdata}
\end{figure}

To create synthetic images related to group emotion, the generator needs photographs of faces showing different emotions. For this, two facial expression databases are available: FACES~\cite{ebner2010faces} and KDEF~\cite{goeleven2008karolinska}.
These datasets contain  frontal facial photos of young, middle-aged, and old people that bring an essential variety to the image collection. As one can see in Figure~\ref{fig:Data_transformations}, the FACES image contains the entire head, the neck and part of the shoulders. 
In our generation process, these images are mixed with a cropped version of the KDEF frontal views\footnote{Downloaded from \href{https://www.ugent.be/pp/ekgp/en/research/research-groups/panlab/kdef}{https://www.ugent.be/pp/ekgp/en/research/research-groups/panlab/kdef}, last accessed 08/2020.} focused on the face\cite{dawel2017perceived,joseph2020facial}, from the chin to the top of the forehead (see Figure~\ref{fig:datageneration}). 
The emotion annotation in both datasets is available in discrete emotion labels, namely the universal facial expressions plus neutral. Due to the presence of the background color on the face images, before overlaying them on the background image, the algorithm gets rid of it using color filtering with OpenCV~\cite{howse2013opencv}.

To enrich the synthetic corpus, data augmentation applies to face images. Wang and al.~\cite{wang2019survey} collected an extensive number of state-of-the-art approaches. From the proposed set of augmentation methods, we identified the most suitable, i.e. the augmentations that distort the initial data in a non-destructive way. Geometric and photometric data transformations are considered. The first type includes transformations that alter the pixels' position in the original image, and the second type encompasses changes in the RGB channels. Figure \ref{fig:Data_transformations} shows examples of the retain photometric and geometric transformations used for face augmentation.

\subsubsection{Validation}\label{sec:validation}
After finalizing the generation process, the first stage was to validate the potential of this data as a training resource. Indeed, before using the synthetic corpus in the challenge system, we had to verify that it is suitable to learn group emotions only with global features. We conducted several tests. As a starting point, we defined a two-class experiment (happiness or not) with only one face within the images. We implemented a model similar to Smile-CNN proposed by Chen \& al.~\cite{chen2017smile}. The authors presented a deep learning architecture focused on detecting one of the most frequent human emotions \cite{trampe2015emotions}, joy (aka happiness in corpora's labels). This architecture intends to detect smiles under both laboratory-controlled and wild data, on single face images. This  straightforward architecture was designed to work on rather small face images ($64\times64$), not on faces in context. The result was equivalent to random.

We decided to evaluate deeper neural networks. Even complexifying the task with up to 9 faces on the image, modified versions of ResNet~18~\cite{he2016deep} and Inception V3~\cite{szegedy2016rethinking} achieved respectively 83.4\% and 88.5\% of accuracy. These results validate that the synthetic dataset can serve as training corpus in group emotion recognition and confirm the potentiality to detect emotions from the whole image with multiple faces. We started our work on the EmotiW Challenge from this point.

\section{Experiments}

This section describes the implementation details, the evaluation of different approaches, the description of the proposed model and the challenge results. For all our experiments, we used PyTorch~\cite{paszke2017automatic} as a deep learning framework. We compared different models by training them using VGAF and our synthetic datasets, validating them on the VGAF validation set.
The challenge organizers were in charge of computing challenge results.

\subsection{Selection of underlying architecture}

Starting from validation results (section~\ref{sec:validation}), we decided to explore exhaustively different state-of-the-art neural network architectures. Indeed, for some computer vision tasks, the advantages and drawbacks of these architectures have been explored. As far as we know, it is not the case for group-level emotion recognition from images. The accuracies of 13 widely employed deep learning architectures were compared on a challenge validation set. Results using these architectures as backbone, modifying them to provide a 3-class classification are presented in Table~\ref{tab:comparison}. Each architecture was modified by changing the number of neurons in the last fully-connected layer to 3, as we have a ternary classification task. All of the considered models were pretrained on the 1000-class ImageNet dataset~\cite{deng2009imagenet}, which contains more than one million high-resolution images. Since these images are significantly different from the challenge's task, these models were then fine-tuned using the VGAF training dataset.

The computed results are quite surprising. First, unlike first results on binary classification (section~\ref{sec:validation}), Inception~V3 has a lower accuracy than ResNet18. Second, the deeper architectures have not the best accuracies.
Among all of the evaluated algorithms, VGG-19 showed the best performance (52.36\% of accuracy). Therefore, this architecture is the basis of the proposed model. The implementation details, as well as challenge results, are described in the next sections.

\begin{table}[ht!]
\centering
\parbox{0.6\linewidth}{
\centering
\label{tab:comparison}
\begin{tabular}{@{}lcccccc@{}}
\bottomrule
\multicolumn{1}{c||}{\textbf{Underlying architecture}}              & Validation accuracy \\  \midrule\midrule
\multicolumn{1}{l||}{Wide ResNet50 \cite{zagoruyko2016wide} } & 43.60\% \\
\multicolumn{1}{l||}{ResNet34 \cite{he2016deep}} & 43.60\% \\
\multicolumn{1}{l||}{MobileNet v2 \cite{sandler2018mobilenetv2}} & 45.04\% \\
\multicolumn{1}{l||}{DenseNet \cite{huang2017densely}} &  45.95\% \\
\multicolumn{1}{l||}{ResNeXt  \cite{xie2017aggregated}} & 47.39\% \\
\multicolumn{1}{l||}{Inception V3 \cite{szegedy2016rethinking}} & 47.91\% \\
\multicolumn{1}{l||}{ResNet50 \cite{he2016deep}} & 48.17\% \\
\multicolumn{1}{l||}{GoogLeNet  \cite{szegedy2015going}} & 48.43\% \\
\multicolumn{1}{l||}{VGG-11 \cite{simonyan2014very}} &  48.69\% \\
\multicolumn{1}{l||}{ResNet18 \cite{he2016deep}} & 49.22\% \\
\multicolumn{1}{l||}{VGG-16 \cite{simonyan2014very}} &  50.91\% \\
\multicolumn{1}{l||}{SqueezeNet \cite{iandola2016squeezenet}} &  51.44\% \\
\multicolumn{1}{l||}{\textbf{VGG19} \cite{simonyan2014very}} &  \textbf{52.36\%} \\
\bottomrule
\end{tabular}
\vspace{0.5em}
\caption{Performance of different network architectures on the VGAF validation set in order of accuracy}
}
\end{table}

\subsection{Challenge model}

This section presents the optimization of the challenge model, its official accuracy and ranking of the team in the Audio-video Group Emotion Recognition sub-challenge.

\subsubsection{Model optimization}
For the challenge, starting from the initial VGG19 model, we further explored several options to improve the model accuracy:
\begin{itemize}
     \item Train the model from scratch, allowing the network to learn more specific deep features, closer to the task;
    \item Vary the size of linear layers;
    \item Add complementary linear layers;
    \item Fine-tune only last layer of our model;
    \item Combine the above options.
\end{itemize}

\begin{figure*}
\centering
\includegraphics[width =0.9\linewidth]{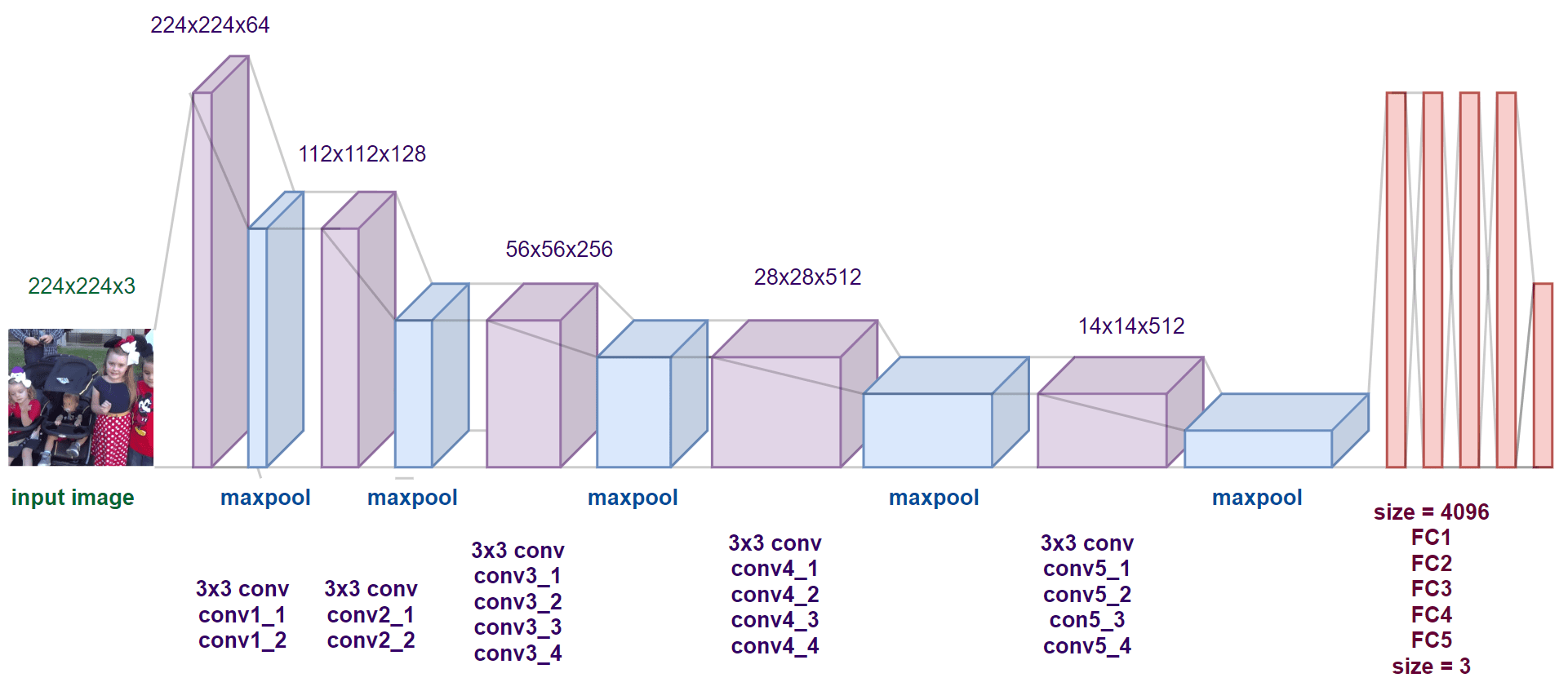}
\caption{Scheme of the modified VGG-19 architecture. The scheme is a sequence of different layers: convolutional, maximum pool, dropout and fully connected layers, used in the architecture.}
\label{fig:vgg19}
\end{figure*}

Comparing all of the aforementioned options, the best performance on the validation set is observed by implementing the following procedure on the model presented in Figure~\ref{fig:vgg19}. First, some fully-connected layers, dropout layers and ReLU activation functions are added at the end of the network, finishing by a ternary tensor. As training from scratch does not provide improvement, the model is first trained on ImageNet. Using these weights, the model is then fine-tuned on both synthetic and VGAF corpora. For each epoch, we generated 10k images using our data generation algorithm and sampled 10k frames from VGAF videos, each frame having an equal probability for being selected. For all images data transformation is performed: arbitrary $\pm 10^{\circ}$ rotation, horizontal flip, random scaling and cropped $224\times224$ regions for frames from the VGAF dataset.
After 10 epochs, the model is trained only with the VGAF dataset using 20k frames for each epoch. 
The training parameters can be summarized as follows. The learning rate equals 0.001. The training script uses stochastic gradient descent with 0.9 momentum as optimizer, and cross-entropy as loss function.

To obtain video-level labels from frame-level classification, we applied two techniques: score averaging and score accumulation. The best results were observed by predicting labels for all frames and then averaging the scores. The final system has a 57.18\% accuracy on the validation set.

\begin{table}
\parbox{.45\linewidth}{
\centering
  \label{tab:metrics}
  \begin{tabular}{cccc}
    \toprule
    \textbf{Class}     & \textbf{Precision}  &\textbf{Recall}    & \textbf{F1-score} \\
    \midrule
    Neutral           & 0.40 &  0.62 & 0.48\\
Positive           & 0.60 &  0.62 & 0.61\\
Negative         & 0.80 &  0.50 & 0.61\\  
Mean value          & 0.60 &  0.58 & 0.57\\
  \bottomrule
\end{tabular}
\vspace{0.5em}
  \caption{Precision, Recall and F1-score for Neutral, Positive and Negative Classes on the validation set}
}
\hfill
\parbox{.45\linewidth}{
\centering
  \label{tab:submission}
  \begin{tabular}{ccl}
    \toprule
    \textbf{Class}              & \textbf{Test accuracy (\%)}\\
    \midrule
Positive & 59.9078\\
Neutral & 46.6019\\
Negative & 75.2174\\

\midrule
All classes & 59.1270\\

  \bottomrule
\end{tabular}
\vspace{0.5em}
  \caption{Accuracy for our best challenge submission.}
\vspace{0.5em}~
}
\end{table}

\subsubsection{Challenge results}
The challenge organizers allowed to make a total of five submissions to evaluate the method on the test set. The team made three attempts, and managed to achieve an accuracy of 59.127\%, reaching the eleventh place out of nineteenth. The best team achieved 76.85\%, potentially with a multimodal approach. 
One can consider that the achieved accuracy is reasonable, given the small size of the training set, the wild data conditions and the fact that our team employed a frugal and unimodal approach with a fairly basic temporal integration.

\section{Discussion}

\begin{figure}[H]
\centering
\includegraphics[width=0.6\linewidth]{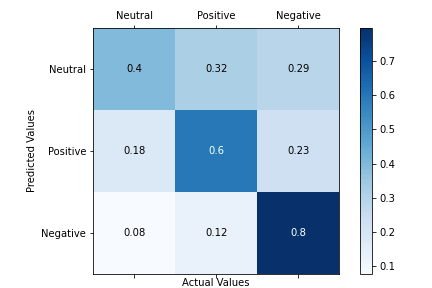}
\caption{Normalized confusion matrix for the validation set for 3-class classification task for the final model. It shows the accuracy of a classification. It could be seen, that the best prediction accuracy for Negative and Positive classes.}
\label{fig:confusionmatrix}
\end{figure}

In order to go deeper into the results, we did several analyses. First, the class-wise accuracy matrix is shown in Table \ref{tab:submission}.
As can be seen from this table, the model shows better performance for the Positive and Negative classes  on the testing set. As expected, the neutral class is the most difficult to classify. To analyze further the misclassifications, information is provided by the confusion matrix on the VGAF validation set (Figure \ref{fig:confusionmatrix}) and in Table~\ref{tab:metrics} that contain various metrics for each of the predicted classes. Similarly to the test set, the Negative class has the highest performance, followed by Positive and Neutral. Almost 60\% of neutral videos have been misclassified, \textasciitilde{}40\% for positive and only \textasciitilde{}20\% for negative videos. As shown in the table, the trained model makes the smallest number of mistakes in predicting the class 'Negative'. It is worth noting that the 'Neutral' class is hard to classify correctly. The recall is low for all classes. The F1-score for Positive and Negative classes are equivalent, while errors differ. The Negative class has the worst recall.
These results are consistent with expectations, and it provides a clear way to improve the performance of our unimodal system, addressing classification behavior with regard to the different classes.

\begin{figure}[H]
    \centering\hfill
    \includegraphics[width=0.3\textwidth]{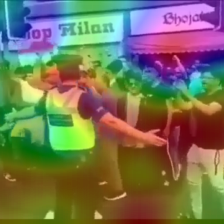}\hfill
    \includegraphics[width=0.3\textwidth]{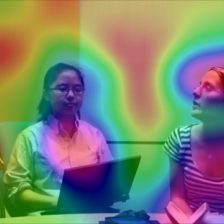}\hfill
    \includegraphics[width=0.3\textwidth]{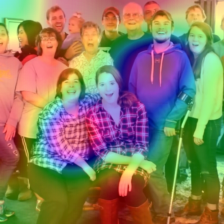}\hfill
    \caption{Negative, Neutral and Positive visualization using Score-cam algorithm~\protect\cite{wang2019scorecam}.}
    \label{fig:score-cam}
\end{figure}

Hypotheses in our methodology are two-fold. The first hypothesis is that using the image at a glance, the neural network will learn to focus on people. Moreover, using our synthetic dataset, the expectation was that it would mainly consider people's faces. The second hypothesis is that the network will ignore the background. To evaluate these aspects, we employed Class Activation Map (CAM) techniques to highlight the contribution of each part of the image to the current class classification. More precisely, we use Score-CAM \cite{wang2019scorecam,uozbulak_pytorch_vis_2019} to compute images shown in Figure~\ref{fig:score-cam}. 

These three images illustrate Score-CAM results on Negative, Neutral and Positive images. Red parts of the heatmap represent less important pixels, while green, blue and purple highlight more and more relevant ones. Looking at these representations, one can conclude that our preliminary hypotheses are not fulfilled. Among these three examples, the best heatmap is for the neutral class. The network actually focuses on both people, ignoring the background. In contrast, in the Negative heatmap, the important pixels are located in storefronts while the network overlooks most people. Last, the positive heatmap reflects that the network concentrates on some people but completely ignores others as if they belonged to the background. Unexpectedly, the neural network does not focus on all people nor their faces. It also does not totally ignore the background pixels. We need to further explore these visualizations to better understand what is relevant and modify the network training accordingly. Nevertheless, these visualizations provide a bag of pixels, but not the underlying visual features (color, shape, texture...). More investigations have to be conducted in order to identify the substance of these features.

\section{Conclusion \& future work}

This paper proposes a deep learning model for group emotion detection using a privacy-safe non-individual approach. The main novelty of this research is the frugal modeling, i.e. processing only global image cues instead of including more and more individual-based features, as well as mixing available datasets with a dedicated synthetic one. We compared thirteen state-of-the-art architectures and created a pure unimodal model able to recognize the mood of groups of people. The final architecture is based on the VGG-19 model with the addition of several fully-connected layers, dropout layers and ReLU activation functions. The experiments showed the promising result of 59.13\% on the EmotiW Challenge test set, in a challenging 3-class classification task \textit{in the wild}.

In the future, some improvements need to be investigated. First, several methods can be used to enhance our synthetic corpus generation: face swapping or replacement in available datasets, facial expression generation \cite{xu2017generative,wang2019u} or even more complex methods targeting  body language.
The way our proposal tackles temporal classification is relatively basic and could be improved. This could be done using Recurrent Neural Networks (RNN) or Long Short-Term Memory (LSTM) networks, for instance. Another way of improvement relies on an in-depth analysis of the proposed network to identify possible enhancements in the network architecture or in the training process. Finally, as most other participants of the challenge did, including the audio stream will be obviously of interest.

To conclude, this research on a unimodal privacy-safe non-individual approach to group-level emotion recognition is promising. The results reinforce the team's ambition to identify classroom ambiance applying ethical rules prohibiting individual computation. Within the context of the \teachinglab{} project, the final goal of such a perception system is to provide clues to teacher trainers to help young teachers improve their pedagogical skills.

\section{Acknowledgments}
This work has been partially supported by the Idex Formation (PIA~2, 15-IDEX-0002) and the ANR 3IA~MIAI@Grenoble Alpes (ANR-19-P3IA-0003) project of the French National Research Agency.

\bibliographystyle{ieeetr}
\balance
\bibliography{bibliography}

\end{document}